\documentclass[letterpaper, 10 pt, conference]{ieeeconf}

\IEEEoverridecommandlockouts                            

\usepackage[english]{babel}
\usepackage{amsmath} 
\usepackage{amssymb}  
\usepackage{graphicx}
\usepackage{subfigure}
\usepackage{caption}
\usepackage{multirow}
\usepackage{amsfonts}
\usepackage{hyperref}
\usepackage{tabularx}
\usepackage[noend]{algpseudocode}
\usepackage{algorithm}
\usepackage{bm}
\usepackage{breakurl}
\usepackage{enumerate}
\usepackage{verbatim}
\usepackage{float}
\usepackage{siunitx}
\usepackage{mdframed}
\usepackage{adjustbox}
\usepackage{cleveref}
\usepackage{authblk}
\usepackage{color}
\usepackage{soul}
\usepackage{cite}

\usepackage{booktabs}  

\title{\LARGE \bf
Towards Autonomous Crop Monitoring: \\
Inserting Sensors in Cluttered Environments 
}

\author{
Moonyoung Lee$^{1}$, Aaron Berger $^{2}$, Dominic Guri$^{1}$, Kevin Zhang$^{1}$, \\ Lisa Coffee$^{3}$, George Kantor$^{1}$, Oliver Kroemer$^{1}$
\thanks{$^{1}$Carnegie Mellon University Robotics Institute,  
        \texttt{\{moonyoul, dguri, klz1, kantor,  okroemer\}@cs.cmu.edu}}
\thanks{$^{2}$Harvard University,
        \texttt{aaronberger@college.harvard.edu}}
\thanks{$^{3}$Iowa State University,
        \texttt{lmcoffee@iastate.edu}}
}

\begin{document}
\maketitle
\thispagestyle{empty}
\pagestyle{empty}

\definecolor{navy}{RGB}{0,0,128}


\begin{abstract}

We present a contact-based phenotyping robot platform that can autonomously insert nitrate sensors into cornstalks to proactively monitor macronutrient levels in crops.
This task is challenging because inserting such sensors requires sub-centimeter precision in an environment which contains high levels of clutter, lighting variation, and occlusion.
To address these challenges, we develop a robust perception-action pipeline to detect and grasp stalks, and create a custom robot gripper which mechanically aligns the sensor before inserting it into the stalk. 
Through experimental validation on 48 unique stalks in a cornfield in Iowa, we demonstrate our platform's capability of detecting a stalk with 94\% success, grasping a stalk with 90\% success, and inserting a sensor with 60\%  success. 
In addition to developing an autonomous phenotyping research platform, we share key challenges and insights obtained from deployment in the field. Our research platform is open-sourced, with additional information available at \textcolor{navy}{\url{https://kantor-lab.github.io/cornbot}}.
\end{abstract}


\section{Introduction}

With the development of artificial intelligence in computer vision and robotics, the agricultural sector is poised to implement precision agriculture methods to enhance crop production efficiency and minimize environmental footprint \cite{review_sensors_robots}.
For example, one of the predominant issues faced by farmers is the overuse of fertilizers, which can be alleviated with increased sensing accuracy and the availability of live data \cite{fertilizer_2,fertilizer}. 
Many existing research works focus on automating vision-based crop monitoring and phenotyping \cite{review_drone_detection, disease_detection, pest_robot, leaf_area, survey_phenotyping}, specifically for detecting cornstalks \cite{corn_slam, corn_mpc}.
However, visible plant features are a lagging indicator of nutrient deficiencies; detecting these deficiencies sooner allows farmers to address crop risk quicker, and thus increase crop productivity.
Therefore, our research focuses on contact-based phenotyping, wherein we insert novel nitrate sensors developed at Iowa State University \cite{sensor,sensor_main} into cornstalks to monitor nitrogen concentration (a useful indicator of plant nutrient health), allowing agronomists to proactively address crop deficiencies.

\begin{figure}[t]
    \vspace{-2pt}
    \centering
    \includegraphics[width=1\linewidth]{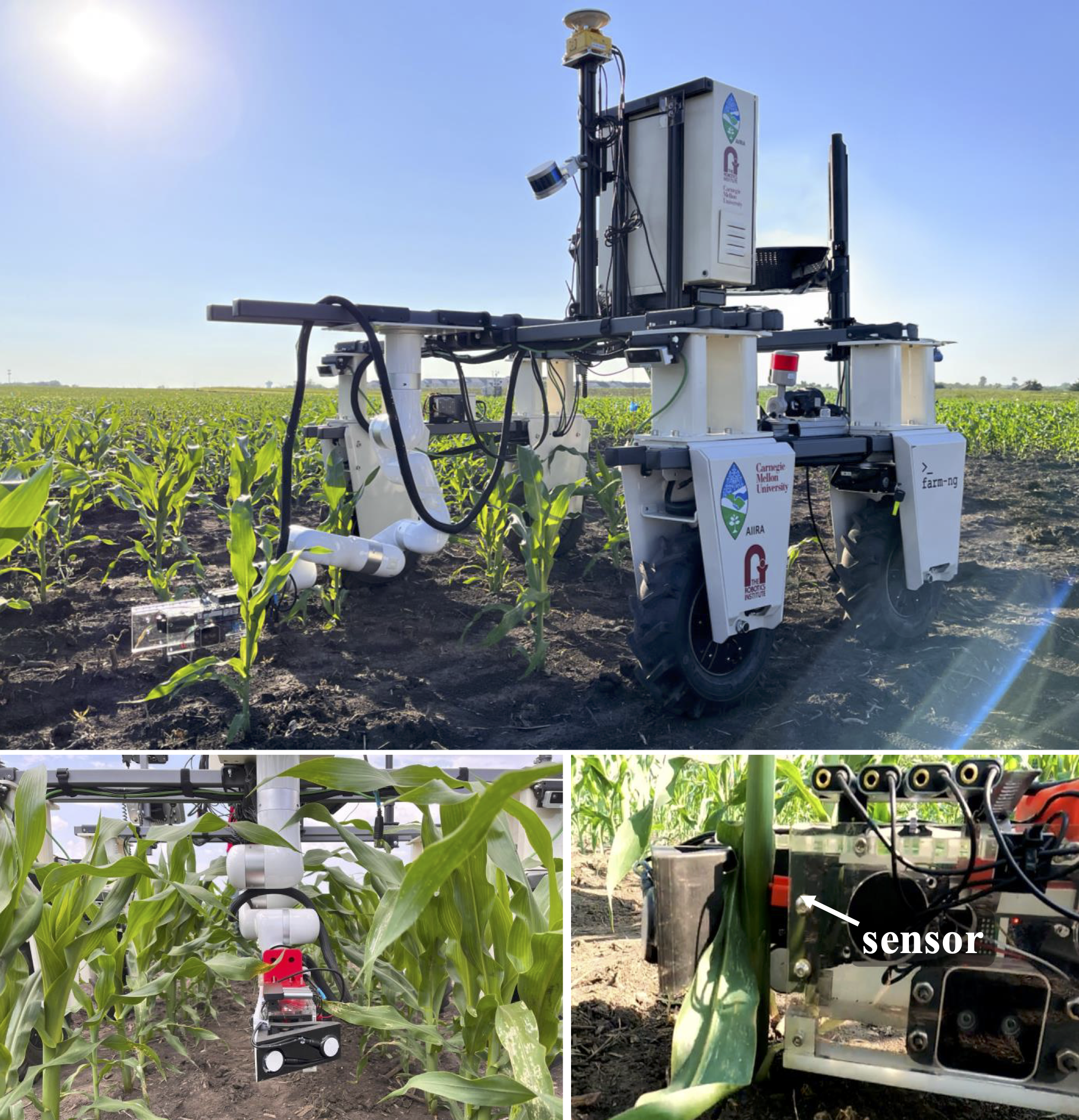}
    \caption{Robot inserting sensors into cornstalks to monitor plant nitrate concentration in Curtiss Farm, Iowa. Our custom gripper utilizes contacts in a clutter-rich environment to precisely align with stalks and insert sensors.}
    \label{fig:cover_pic}
    \vspace{-15pt}
\end{figure}

\begin{figure*}[!ht]
    \centering
    \includegraphics[width=0.98\linewidth]{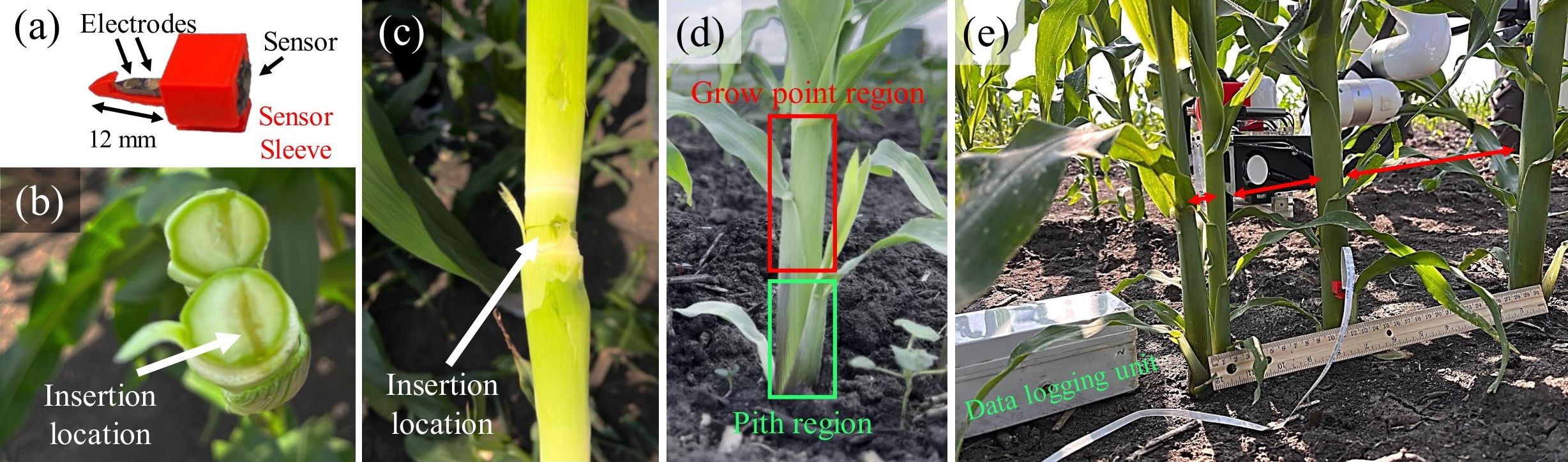}
    \caption{Overview of the task: (a) Sensor to convert nitrate to voltage and the sensor sleeve for protection when penetrating into stalks; (b) Cross-section view of a stalk after a successful insertion; (c) View of the insertion location after a deployed sensor has been manually removed; (d) View of a stalk showing the pith region, a solid node near the root, and the grow-point region composed of layers of leaves; (e) Cluttered field environment with varying stalk spacing, occlusion from leaves, and uneven ground terrain. The deployed sensor is wired to the data logging unit to store data over the plant's life cycle.}
    \label{fig:corn_diagram}
    \vspace{-15pt}
\end{figure*}
While less abundant than vision-based phenotyping methods, there are several contact-based phenotyping studies that interactively measure plant traits using contact-based sensors. Examples include robots with custom leaf-grasping grippers which measure chlorophyll content with a SPAD meter \cite{2012alenyaRoboticLeafProbingb} or spectral reflectance with a spectrometer \cite{2019atefiVivoHumanlikeRobotica}.  Sensor penetration through leaves has also been demonstrated in \cite{2017baoRobotic3DPlant,shahDevelopmentMobileRobotic}, which use a needle-like fluorometer to detect chlorophyll fluorescence emission. In addition to interacting with leaves, there have been efforts to grasp cornstalks to measure diameter \cite{grasp_diameter}. 
Our work is most similar to contact-based phenotyping methods deployed on sorghum plants, where a force-gauge sensor is probed into a stalk to measure strength \cite{robotanist_cmu, grasping_sorghum}. 
However, unlike reusing a single probe for multiple measurements, our task involves deploying sensors in multiple plants, which assists a farmer in taking action to address deficiencies in a large area of a field.

Manually installing these nitrate sensors with corresponding data logging units across a farm is a laborious process that is difficult to scale \cite{review_phenotyping_robot, review_robot_plant_phenotyping}.
Thus, we present a robotic research platform capable of autonomously inserting nitrate sensors and deploying corresponding data logging units in the field. 
We believe our presented work can benefit the agricultural robotics community by identifying key challenges in deploying a robot for contact-based phenotyping. By sharing our design decisions and valuable insights obtained in the field, we hope to contribute to building a fully autonomous phenotyping research platform that can support farmers.

To summarize, the contributions of this paper are: 
\begin{itemize}
    \item A platform capable of autonomously deploying nitrate sensors into stalks for contact-based phenotyping
    \item Extensive evaluation of our robotic platform on 48 sensor insertion trials at Curtiss Farm, Iowa
    \item A dataset of 7600 annotated cornstalks for segmentation tasks
\end{itemize} 
    %

 %

\section{Challenges of Sensor Insertion in Corn}~\label{sec:challenges} 
\vspace{-10pt}

In contrast to robotics experiments conducted in a controlled indoor lab environment, agricultural robotics is challenging due to extreme variations in field conditions and limited accessibility to test a system due to the narrow time frame of seasonal crops. For the task of cornstalk sensor insertion, we specifically discuss the challenges that arise in near-ground precise manipulation tasks and the perception difficulties when operating in a cluttered, outdoor environment.

\subsection{Contact Interaction Challenges}
One of the main challenges is the sub-centimeter precision required to insert a small sensor (whose probe is 12x3x2 mm in dimension, shown in Fig. \ref{fig:corn_diagram}a) for accurate data logging. The stalk diameter and probe length are of similar magnitude, so it is important to precisely insert sensors near the vertical center of the stalk. If the sensor is instead inserted near the edge, the insertion may not be deep enough to fully immerse the sensor's two electrode pads into the stalk, leading to faulty sensor readings. Examples of successful insertions are shown in Fig. \ref{fig:corn_diagram} (b,c,e). 

The variation of stalk shape and diameter observed in the field make this required precision difficult. For example, while some stalks have circular cross-sections, a large portion have elliptical cross-sections, making the direction of insertion important. For the purpose of monitoring crop nitrate concentration, it is ideal to insert sensors in corn plants in the V4-V8 stages of the plant's life cycle \cite{sensor_main}. The stage of a corn plant is measured by the number of leaf collars: V4-V8 plants have four to eight leaf collars, a stalk diameter between 12-40 mm, and are typically grown in 14-28 days. 

An additional requirement for the sensor to properly monitor nitrate concentration is to be placed in the pith region of the stalk, referred as the first node typically 2-10 cm above the ground (Fig. \ref{fig:corn_diagram}d). The pith region does not structurally change over the plant's growth cycle, allowing for consistent measurement from an embedded sensor. In the grow-point region, however, the stalk contains layers of leaves that continuously grow over the life cycle and thus force out the embedded sensor from within the stalk. This requirement to operate in close proximity to the soil poses a challenge: the terrain of a cornfield is highly uneven, and ditches and mounds may collide with the robot arm.

Another challenge is the high variation in corn plant spacing, as shown in Fig. \ref{fig:corn_diagram}e. Even when corn seeds are planted with a predetermined density on a seeder machine, tillers (reproductive shoots growing upward from the same plant) and other growing conditions lead to narrower or wider gaps between stalks.

\subsection{Perception Challenges}
These described variations in stalk shape and spacing also make perception difficult, as heuristics like expected stalk spacing and maturity are inaccurate.
Varying lighting conditions depending on time of day also pose a challenge: as the robot traverses through corn plant rows, the camera switches between facing east and west. In the morning and evening, these viewpoints introduce problems with sun flare or severe overcast shadows from the stalks and the robot itself.
Another challenge arises from the frequent occlusion of stalks from nearby leaves, as shown in Fig \ref{fig:cover_pic}. Lastly, targeting the near-ground pith region is difficult as highly uneven terrain make vision-based height predictions inaccurate—ground plane detection methods like RANSAC and image segmentation were tested, but did not accurately identify ground height around the corn plants.

%

\section{Robot Platform Overview}~\label{sec:overview}
\vspace{-10pt}

\begin{figure}[t]
    \centering
    \includegraphics[width=1\linewidth]{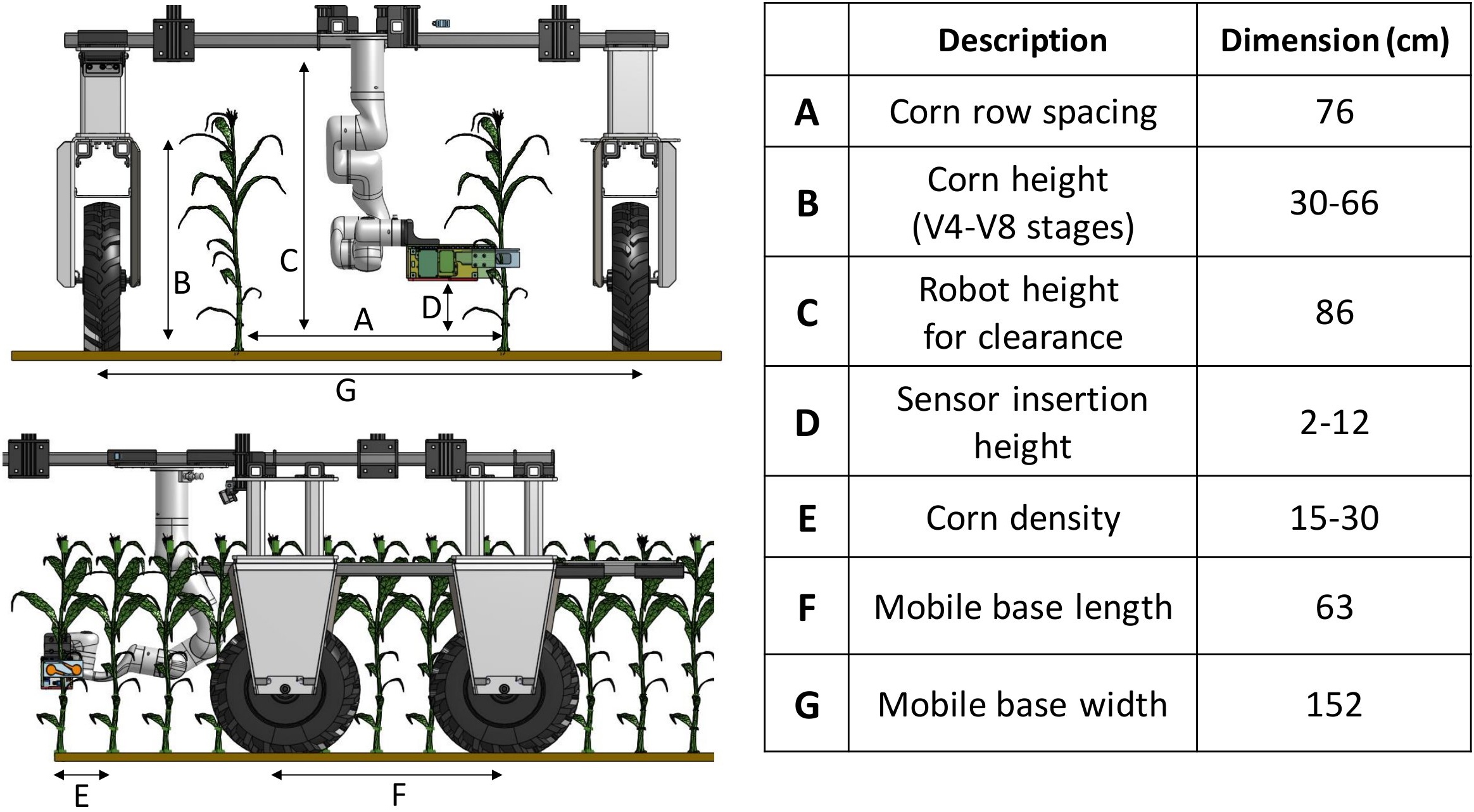}
    \caption{Cornfield specification, which determines the key dimensions of the robot platform.}
    \label{fig:robot_dimension}
    \vspace{-15pt}
\end{figure}

The robot platform consists of a six degree-of-freedom xArm robot attached to a four-wheel-drive mobile platform with a custom end-effector for the sensor insertion task.
The mobile base is built upon the commercially available Amiga platform from farm-ng\cite{amiga} which allows for quick hardware adaptations to the specialized agriculture task at hand. Our mobile base is modified by adjusting the bar extrusion for width and height according to the field specifications at Curtiss Farm, Iowa, as shown in Fig. \ref{fig:robot_dimension}. The width of the mobile base (between left and right wheel centers) is set at 1.5 m to accommodate the 0.75 m corn row spacing. The height of the bottom of the robot platform is set according to the expected corn plant height at the V4-V8 stages, such that the robot's platform hovers above the two straddled corn plant rows.

All software for the robot runs onboard on the mini-ITX motherboard containing an Intel i9 24-core 3 GHz CPU and an external RTX4070 GPU, enclosed in a weather-resistant box. The platform is powered by two Lithium-ion batteries providing a roughly 3 hour run-time in the field.

\section{Gripper Design }~\label{sec:gripper}
\vspace{-15pt}

\subsection{Design Motivation}\label{sec:design}

\begin{figure}[t]
    \centering
    \includegraphics[width=\linewidth]{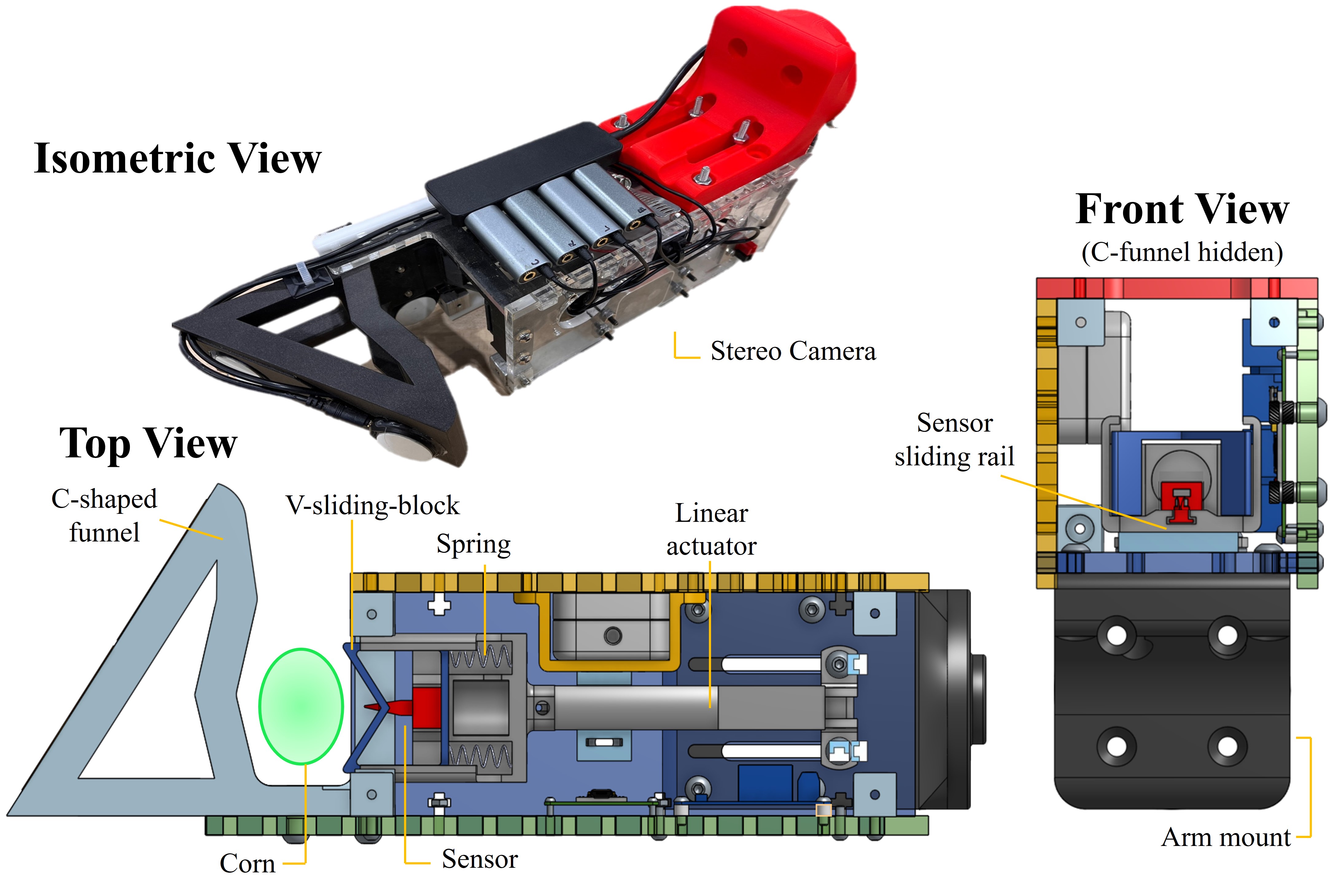}
    \caption{Custom gripper for senor insertion. The V-sliding block is actuated to press against the stalk, compressing the spring and allowing the sensor to protrude out. The funneling edges help center-align the stalk with the sensor.}
    \label{fig:gripper_diagram}
    \vspace{-10pt}
\end{figure}

As discussed in Sec.\ref{sec:challenges}, the main challenges for precise sensor insertion in the field are variation in stalk attributes (shape, diameter, and spacing), clutter, and uneven terrain. While there are visual methods to reduce pose uncertainty such as multi-view pose estimation and visual servoing \cite{visual_servo}, these approaches are less applicable in cluttered environments due to occlusion and collision risks. Instead, to achieve the required precision, we design the gripper system to mechanically induce the alignment of the sensor and corn stalk using funneled edges and spring-loaded compliance. For compactness, the gripper utilizes only one linear actuator to push the sensor into the stalk.

\subsection{Gripper Mechanism}\label{sec:design}
To fit between corn plants and insert sensors near the ground, the gripper is compact, measuring 254 $\times$ 76 $\times$ 76 mm (L $\times$ W $\times$ H). The gripper contains an Intel D405 stereo camera for stalk detection, a 50 mm stroke linear actuator for linear sensor insertion, a spring-loaded V-sliding-block for compliance, and a rigid C-shaped funnel piece (Fig.~\ref{fig:gripper_diagram}) for centering the stalk for insertion. The linear actuator's force of 90 N and stroke length of 50 mm were empirically determined to best address stalk rigidity and diameter in a plant's V4-V8 stages.

\begin{figure}[t]
    \centering
    \includegraphics[width=\linewidth]{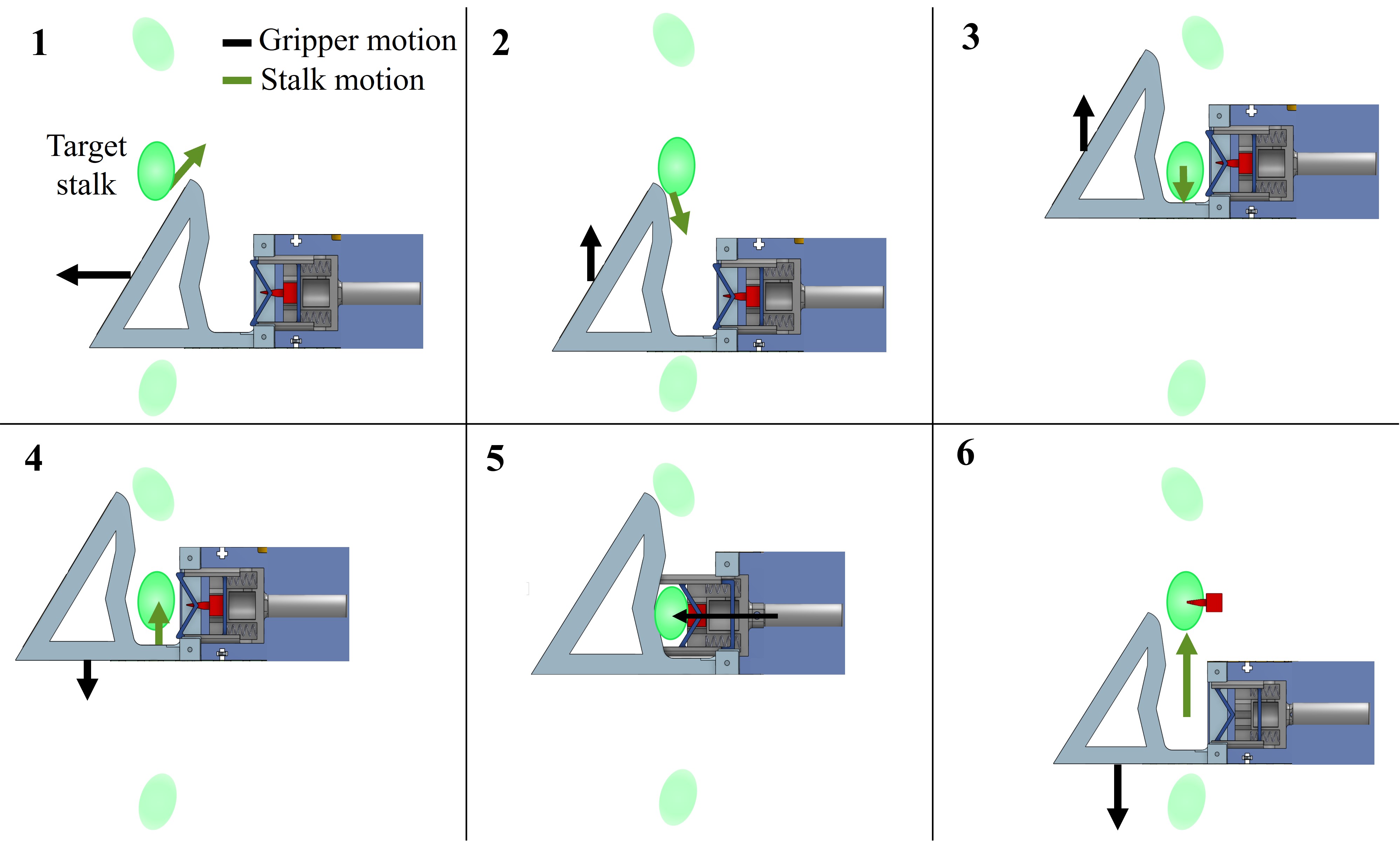}
    \caption{Sequence of contacts with the stalk for sensor alignment.}
    \label{fig:sequence}
    \vspace{-15pt}
\end{figure}

\begin{figure*}[!ht]
    \centering
    \includegraphics[width=0.9\linewidth]{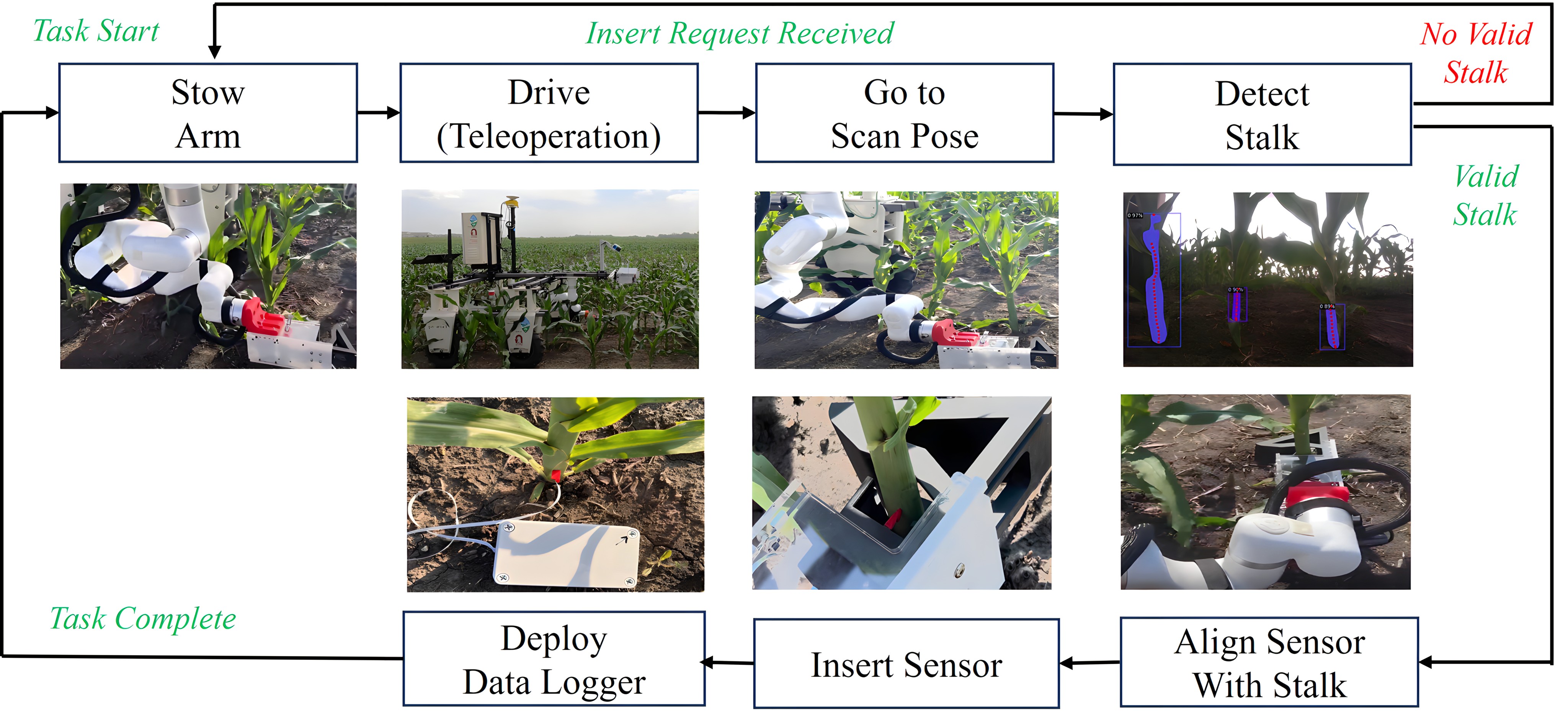}
    \caption{Overview of the sensor insertion sequence. The user manually drives to a general location and initiates the sequence. The xArm scans for a valid stalk, executes an open-loop motion that utilizes mechanical alignment in the gripper, inserts the sensor, deploys the data logging unit, and awaits command in the next region.}
    \label{fig:fsm_diagram}
    \vspace{-15pt}
\end{figure*}

The front end of the C-shaped funnel is designed as a wedge to traverse through corn plant leaves while approaching the targeted stalk. In cases where the gripper collides with the targeted stalk during the approach motion due to pose estimation or motion execution error, the wedge also guides the stalk until it enters the funnel. 
As the robot arm moves laterally to position the stalk inside the funnel, the funneled edge comes in contact with the compliant stalk and guides it to the region precisely in front of the V-sliding-block (Fig. \ref{fig:sequence}). The linear actuator then pushes the V-sliding-block to the wall of the funnel where the stalk is located. Because of the stalk's compliance and roundness, as the V-sliding-block comes in contact with the stalk, the stalk further aligns to the cusps of the V-sliding-block. Since the V-sliding-block is spring loaded, upon contact with the stalk, the block slides backward, exposing the rigid sensor to penetrate the stalk. This spring loading is critical, since the stalk is further aligned by the V-sliding-block before the sensor makes contact. Without the spring loaded compliance, the sensor would often initially prod the stalk off-center, at which point further alignment by the V-sliding-block would not re-align the sensor leading to failed insertions.

\subsection{Sensor Deployment}\label{sec:deploy}

To measure the nutrient content in cornstalks, we utilize the custom nitrate sensor developed by Ali et al\cite{sensor_main}. The sensor relies on the electrochemical reaction from the solution applied on the surface of the printed circuit board (PCB) to generate voltage across the two electrode pads. This solution on the surface of the PCB can wear and tear during the stalk penetration process. To prevent this sensor damage, the sensor is fitted inside a 3D printed sensor sleeve from PLA material, which protects the membrane, as shown in Fig. \ref{fig:corn_diagram}a.


The sensor sleeve is positioned on a T-slot rail on the V-sliding-block, which restricts the sleeve motion laterally, as shown in Fig. \ref{fig:gripper_diagram}. During the actuator extension process, the solid backend of the T-rail holds the sensor in place. During the actuator retraction process, the arrowhead grips to the stalk, providing sufficient force to pull the sensor out from the T-rail. The T-rail tolerance is balanced to provide sufficient friction to hold the sensor in place during motion while allowing the needed tolerance to easily slide in and out during insertion (and reloading a new sensor).


 %

\section{Insertion Motion}~\label{sec:Motion}

\vspace{-15pt}

\subsection{Arm configuration}\label{sec:simulation}

\begin{figure}[t]
    \centering
    \includegraphics[width=\linewidth]{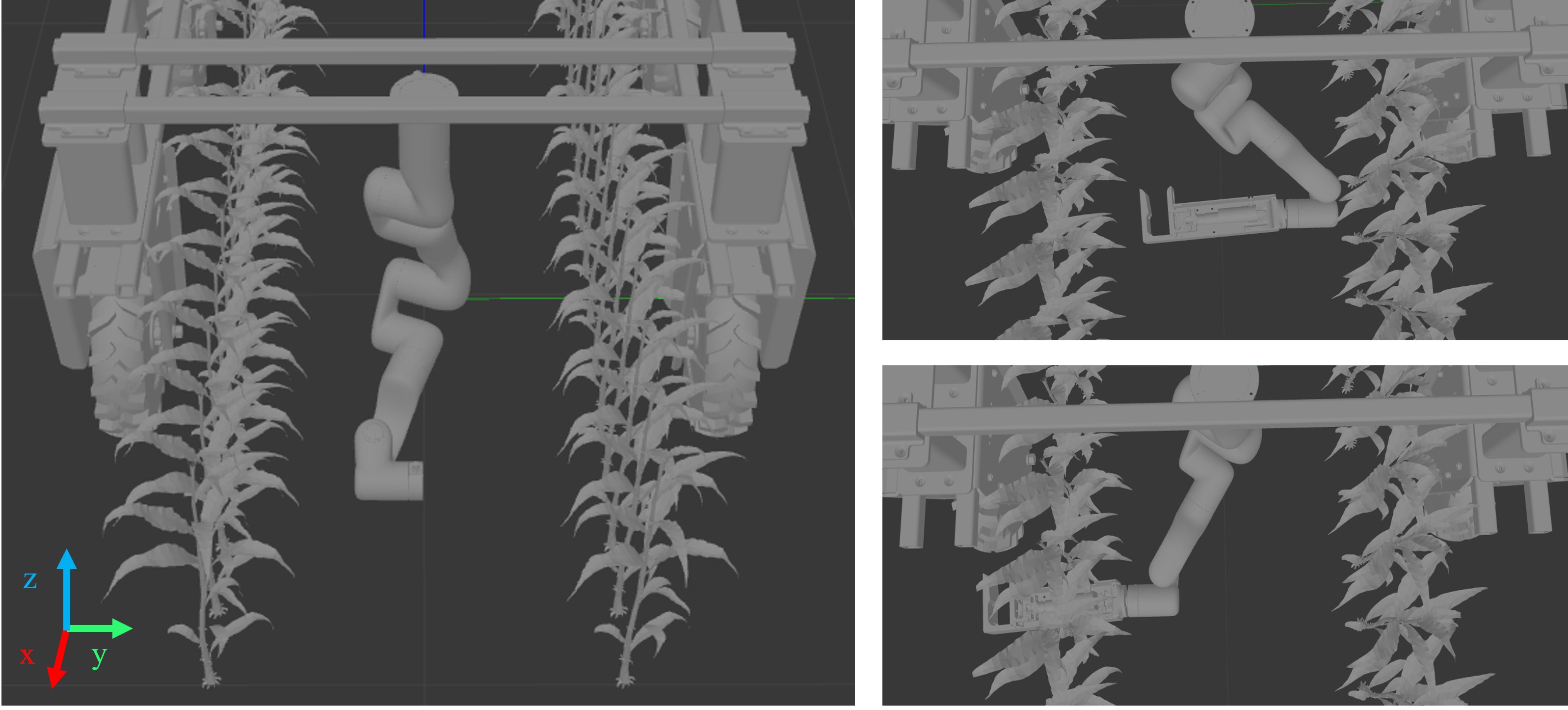}
    \caption{Arm placement validation conducted through simulation by checking the motion range needed for insertion.}
    \label{fig:sim}
    \vspace{-10pt}
\end{figure}

The xArm is mounted in an inverted configuration, aligned to the center of the mobile platform and towards the front, so that stalks on either side can be targeted as shown in Fig. \ref{fig:robot_dimension}, and the arm can be easily accessed for reloading nitrate sensors in testing.
In order to validate that the robot is able to kinematically reach targeted stalks in both rows, motion tests were performed with the robot platform in the Gazebo simulator by moving from the arm's initial position to various possible stalk X,Y,Z positions. As the stalks have no collision properties in simulation, the RRT* trajectory planner provided in MoveIt returns feasible trajectories that allow the arm to reach targeted stalks, as shown in Fig. \ref{fig:sim}.


\subsection{Insertion Motion Sequence}\label{sec:fsm}

The software for the insertion motion sequence is built on a ROS SMACH task-level state machine \cite{smach}. The finite state machine (FSM) enables modular development of each task. Fig. \ref{fig:fsm_diagram} shows the flow diagram of the insertion sequence, including the xArm motion primitives, stalk detection, sensor insertion, and data logger deployment. 

The sequence starts with the robot arm in a stowed position, and the operator teleoperates the mobile base to a general region in the cornfield. Given an insert command, the robot arm moves out of the stowed position to a scan position in order to visually detect cornstalks.
As later discussed in Sec. \ref{sec:method_detect}, a detect request to the detection pipeline returns either a reposition response—in which case the robot halts the motion sequence due to lack of visible stalks—or a stalk insertion position. This insertion position never exceeds the xArm's kinematic workspace. Given a valid stalk pose, the gripper approaches the stalk with a predetermined sequence of actions using a combination of joint and Cartesian space commands. In order to align the stalk with the sensor inside the funnel, the arm makes intentional contact with the stalk to guide it into position and locally align the sensor as discussed in Fig. \ref{fig:sequence}. Once aligned, a serial communication to the Arduino inside the gripper triggers the two-channel relay to extend and retract the linear actuator, ramming the sensor into the stalk. As the actuator retracts, the arrow-hook in the sensor grips into the stalk causing the sensor to slide out of the T-slot rail as it retracts. Finally, to deploy a datalogging unit that stores sensor measurements over the plant growth cycle, a serial communication to a separate Arduino inside the deployment box opens one of the five available datalogging units to be released.

Although the robot is operating in a highly cluttered area, the motion sequence does not do 3D scene mapping to explicitly plan for collision-free trajectories. Instead, the motions are a heuristically determined sequence of open-loop trajectories to align the stalk inside the gripper based on the given insertion pose. Since most contacts occur with compliant leaves and plants, this implementation decision does not lead to critical collisions. By not handling external collision avoidance, the motion sequence is fast and lightweight. Furthermore, in the few instances of collision with the ground, the arm safely stops using current limits on all arm joints.

\subsection{Validation on Mock Cornfield}

\begin{figure}[t]
    \centering
    \includegraphics[width=\linewidth]{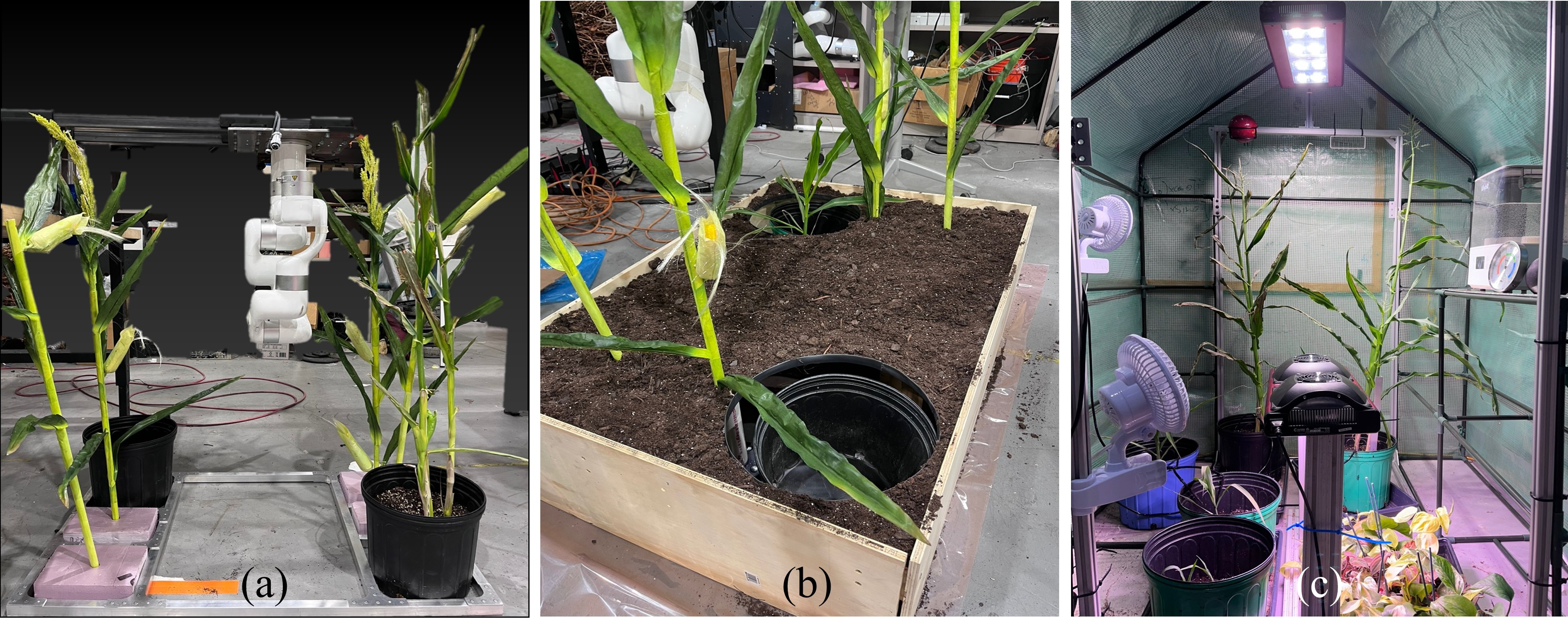}
    \caption{Mock cornfield setup for indoor evaluation: (a) xArm configuration emulated for field condition; (b) Mock field with synthetic and real corn; (c) Greenhouse created to grow corn plant samples for insertion.}
    \label{fig:mock}
    \vspace{-10pt}
\end{figure}
Crucial gripper design decisions, such as the force required to penetrate a cornstalk and optimal dimensions for a funnel, are best determined from real corn plant samples. However, as there are no cornfields available before the summer, cornstalks were grown indoors using a greenhouse.
The mock field setup in the lab (Fig. \ref{fig:mock}) enabled the integration of the arm motion sequence and detection pipeline discussed in Sec \ref{sec:method_detect} prior to field testing. This included validating the in-hand calibration so that the gripper could grasp the targeted stalk given its detected position from the camera.

\section{Visual Detection}~\label{sec:method_detect}

\vspace{-15pt}

Sunlight induces strong interference with common active IR RGB-D cameras. We therefore utilize a stereo-based RGB-D camera on the robot gripper, which provides accurate 3D data at close ranges. The perception pipeline's purpose is to reliably determine a sensor insertion point in 3D space, optimally in the pith region (the first node) of a stalk, which is generally the bottom 2-10 cm of the plant protruding from the ground, based on stalk maturity.
The insertion point's $z$-coordinate is therefore set as a tuned hyperparameter, since the insertion height varies across fields and vision-based ground plane height estimations proved to be inaccurate due to uneven terrain.

The input to the pipeline is a sequence of RGB-D frames and the output is a stalk 3D position relative to the robot. In each frame, 2D masks of stalks are segmented and the best stalk is determined from 3D stalk attributes extracted from the pointcloud. After all frames are processed, the best insertion position is determined by a consensus among frames.


\begin{figure}[t]
    \centering
    \includegraphics[width=\linewidth]{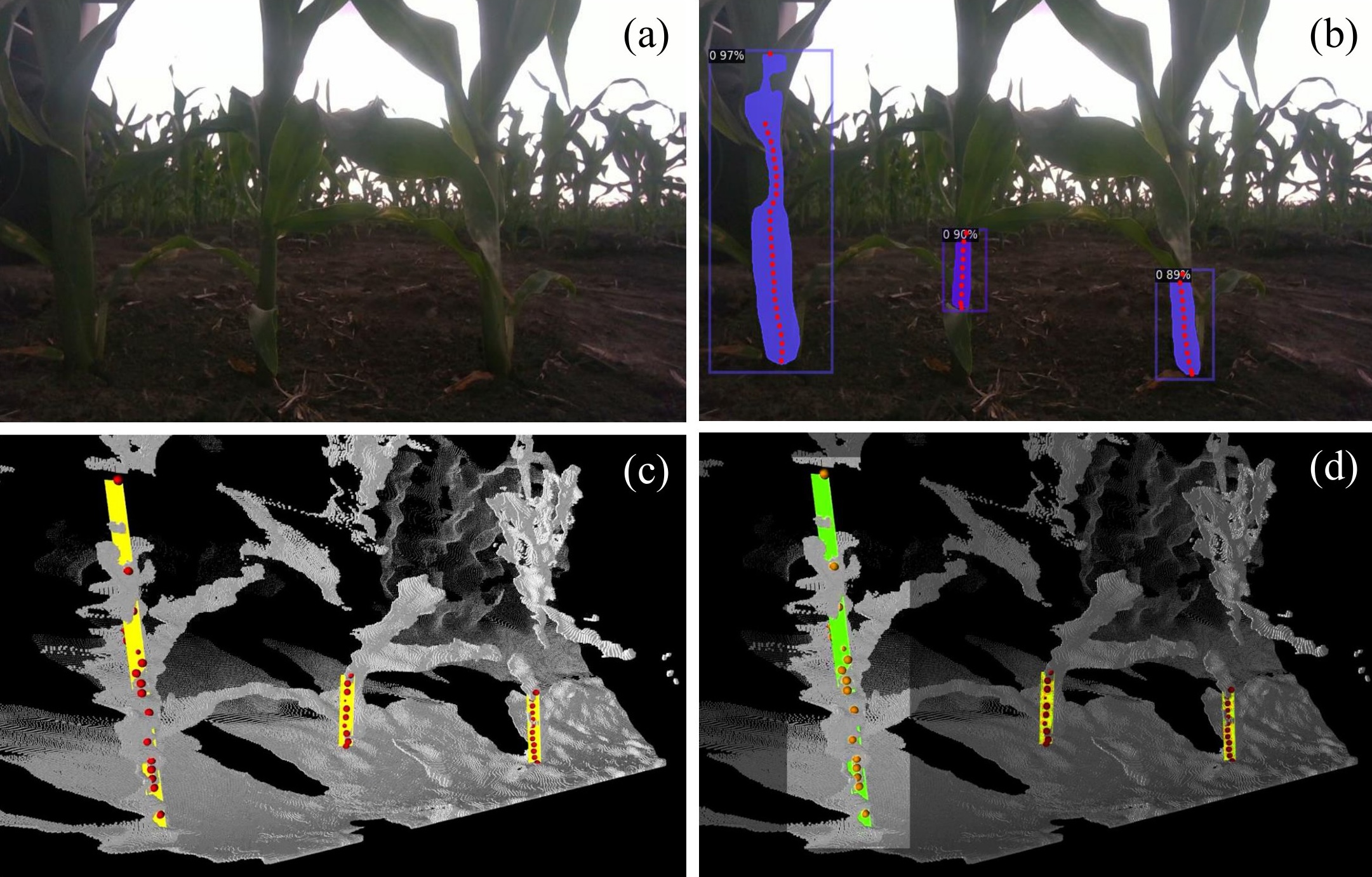}
    \caption{Stalk detection pipeline: (a) Input image from the in-hand camera; (b) Mask segmentation and the extracted 2D feature points; (c) 3D feature points and fitted lines along the stalks; (d) Extraction of the best stalk for insertion based on scoring metrics from mask and stalk features.}
    \label{fig:detection}
    \vspace{-10pt}
\end{figure}

\subsection{Creating the Stalk Detection Dataset}
Variations in lighting, time of day, plant maturity, and field conditions at test-time necessitate a diverse image dataset for training a deep segmentation model. Due to limited access to the cornfield with the robot, the dataset images are collected from videos captured on an iPhone 13 in various regions and sunlight conditions (at 10AM and 5PM) at the Curtiss Farm. During recording, the phone is positioned to emulate the viewpoint of the camera on the robot arm—approximately 15 cm above the ground.

First, one frame per second is taken from the video files while removing extraneous images, resulting in a collection of 2667 RGB images. These images are then manually labeled using Meta's Segment Anything model \cite{SAM}, which efficiently assists the interactive annotation process. Masks are specifically applied to stalks in the foreground, with each mask extending from the bottom of the visible stalk up to the point of occlusion by the plant's leaves. The labeling process resulted in 7681 stalk instance annotations. The dataset is then split randomly into 80\% training, 10\% validation, and 10\% testing data. To our knowledge, this is the largest corn stalk dataset with segmentation annotation (3.5x the size of the dataset used in \cite{corn_slam} and 7.4x that of \cite{stalk_counting}).

\subsection{Stalk Detection}\label{sec:detection}
Given an RGB image as input, the segmentation model outputs 2D masks of all stalks in the foreground. The segmentation method is based on the Mask-R-CNN architecture \cite{He2017} with the ResNet-50 backbone and Feature Pyramid Networks for segmentation. A pretrained Mask R-CNN model is fine-tuned on our training data for 40,000 iterations, with a maximum learning rate of 2.5e-4 and a batch size of 4. Training took 6.7 hours on an NVIDIA GeForce RTX 3060. Quantitative evaluation is discussed in Sec. \ref{sec:detect_eval}.

\subsection{Stalk Pose Estimation}\label{sec:pose_estimation}
The first step in estimating the best insertion pose in a single frame is to determine a set of \textit{feature points} along each stalk from the segmented masks. As shown in Fig. \ref{fig:detection}b, \textit{feature points} are center points along the stalk mask, vertically distanced by 10 pixels between points. Corresponding depth values for each \textit{feature point} are used to project to 3D points. 

However, since these segmented masks may be partially occluded by nearby leaves, it is important to note that the bottom of the mask does not necessarily correspond to the ground. Thus, a line is fitted—using RANSAC followed by Least Squares refinement—along the stalk in 3D space to deduce the insertion height from the ground as shown in Fig. \ref{fig:detection}c. The line is then extended to the ground plane, under the assumption that the robot coordinate's ground plane is at zero height. In comparison with other tested ground plane extraction methods—RANSAC plane detection on the pointcloud, and image segmentation and projection to 3D—this method achieved the most consistent results.
This fitted line, which is now more robust to partial occlusion, is used to determine the insertion point at the hyperparameter $z$-height.

To select the best stalk in which to insert a sensor, the detection pipeline incorporates simple yet effective spatial reasoning to reject candidates that pose difficulties for the robot arm. First, stalks that are outside the manipulator's workspace—based on predetermined boundaries—are rejected. Second, stalks that are positioned too close together, such that the gripper's width prevents it from reaching the targeted stalk, are also eliminated.

The best stalk is chosen from the remaining candidates based on the following heuristic weighting function:
\begin{equation}
    \textrm{best score} = \arg\max(c^2\times w \times \sqrt[3]{h} \times (1 - d))
\end{equation}

where $c$ is the segmentation confidence score ($0 \le c \le 1$), $w$ is the segmentation mask's width, $h$ is the mask's height, and $d$ is the mask's horizontal distance from the center of the image frame. Thus, preference is given to larger stalks with higher confidence which are closer to the center of the image. The powers on the $c$ and $h$ term are selected from empirical testing—most scores ($c$) are close to 1 so squaring better emphasizes differences in score, and the height of a stalk matters relatively less than its other attributes, so the $\frac{1}{3}$ power reduces this attribute's weight. Finally, to reduce the potential effect of noise and faulty segmentation, this process is repeated for five frames and the best stalk among all frames is determined by clustering the stalk lines and selecting the representative from the largest cluster. In cases where no valid stalks are detected, a \textit{reposition} response is issued to indicate moving the robot to an another viewpoint. 
\section{Experimental Results}~\label{sec:experiments} 
\vspace{-10pt}

To evaluate our platform’s ability to insert sensors into cornstalks, we deployed the robot at Curtiss Farm in Iowa during July, in an approximately 100x30 meter region of the cornfield with V8 stalks (about 66 cm tall). First, we evaluate the perception pipeline as a subsystem. Then, we assess the performance of the overall insertion pipeline and analyze failure cases.

\subsection{Stalk Detection Evaluation}\label{sec:detect_eval}

\begin{figure}[t]
    \centering
    \includegraphics[width=\linewidth]{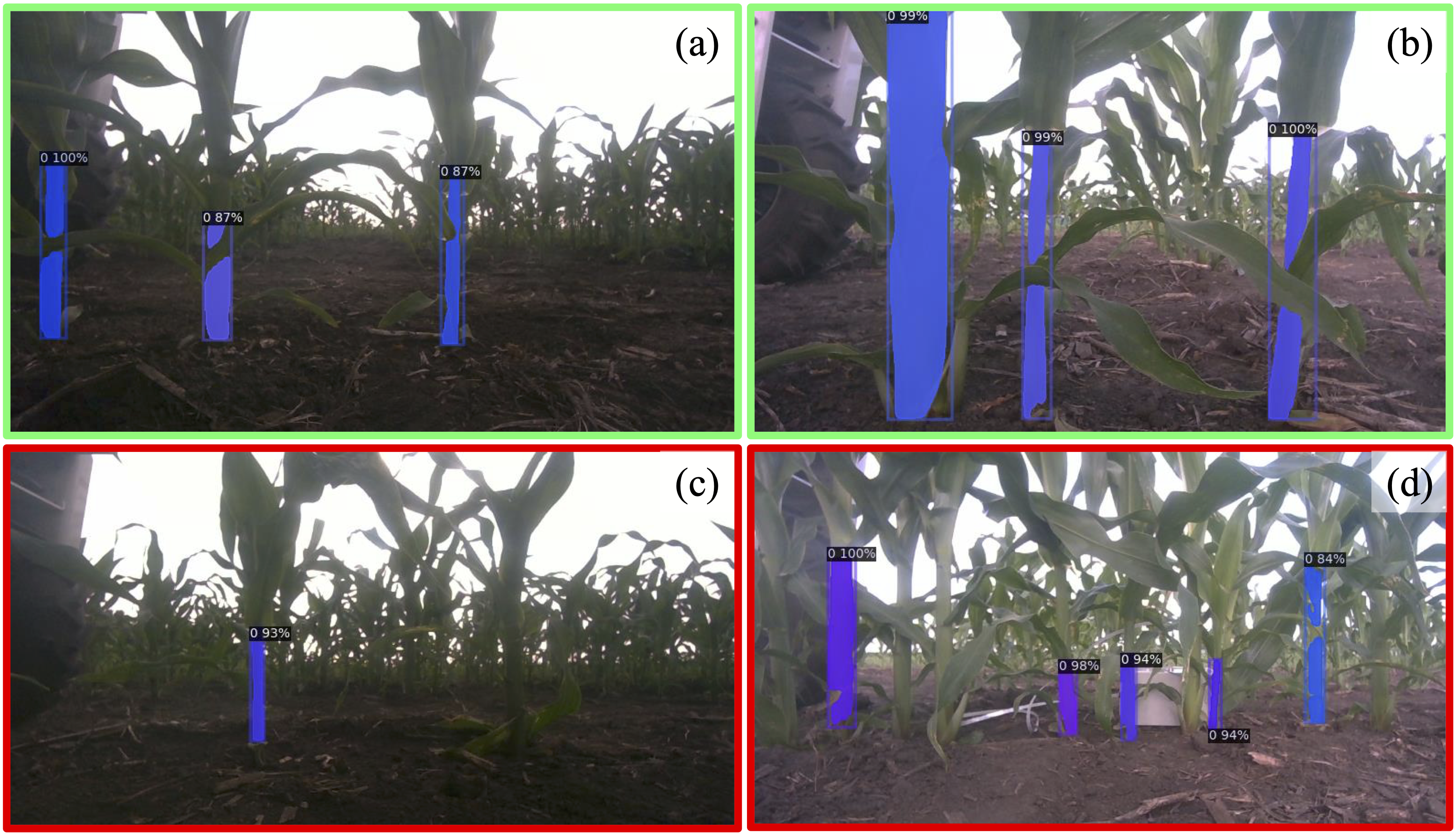}
    \caption{Segmentation success and failure cases: (a) Successfully segmenting stalks around leaf occlusions; (b) Successfully segmenting stalks with highly varying width; (c) missing a stalk due to shadows in poor lighting conditions; (d) Missing multiple stalks due to occlusion.}
    \label{fig:detections}
    \vspace{-10pt}
\end{figure}
We evaluate both the performance of the 2D segmentation model and the 3D position estimation precision. Our 2D segmentation model demonstrates accurate stalk segmentation on unseen images, resulting in 81\% average precision at Intersection over Union (IoU) of 0.5 (and 51\% at IoU 0.75) on the test dataset. Frequent occlusions by leaves (which are often mis-segmented by the model as part of the stalk), along with inconsistencies in where the top of the stalk is defined in the annotated labels, lead to lower average precision especially at high IoU. Fully missed detections are mostly attributed to high amounts of clutter and shadows over the foreground stalks by the robot. Examples of results on challenging scenes are shown in Fig. \ref{fig:detections}.

We evaluate our 3D position estimation performance by comparing the predicted insertion point to a hand-measured ground truth. The ground truth is obtained with a tape measure to determine the forward and lateral distance to the detected stalk from the edge of the camera lens. We disregard the error in height since the height of the insertion point is set as a hyperparameter. Over 100 estimations, we observe consistently robust pose estimations, resulting in mean accuracy of approximately 5 mm. We observe that the estimation accuracy decreases for thinner and further plants.

\subsection{Field Evaluation}\label{sec:field_eval}

\begin{figure}[t]
    \centering
    \includegraphics[width=\linewidth]{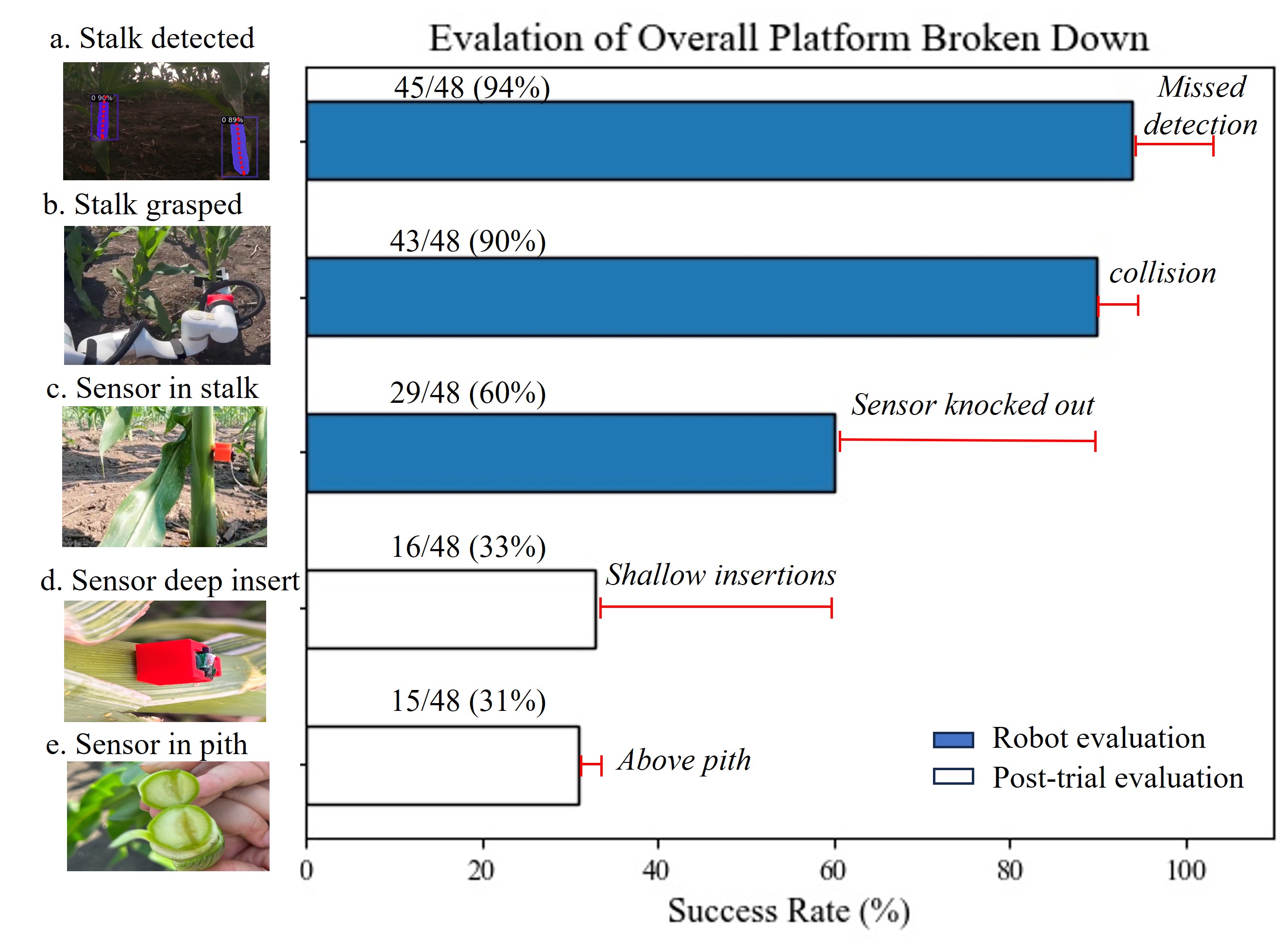}
    \caption{Overall platform evaluation from 48 insertion trials. We measure 5 criteria in increasing complexity for insertion success. }
    \label{fig:eval_fail}
    \vspace{-10pt}
\end{figure}

We conduct insertion trials on 48 unique cornstalks to assess performance, which is 3x more evaluation samples than in a similar work for stalk grasping\cite{grasping_sorghum}. On average, each trial takes two minutes to drive to a new stalk and insert a sensor. The result for each trial is categorized into five criteria for evaluation, as shown in Fig \ref{fig:eval_fail}.

The stalk detection pipeline achieves 94\% success rate, returning 3 missed detections out of 48 trials when valid stalks are within the field of view (a success is defined as returning a reasonable position on a stalk). It is important to note that the cameras used during the training of the segmentation network (iPhone 13) and testing (Realsense D405) are different. Nonetheless, the pretrained Mask R-CNN backbone is robust in identifying the learned features of cornstalks.

The robot successfully grasps 43 stalks out of the 45 detected stalks. As the cornstalks are static targets, the open-loop insertion motion sequence proves sufficient, with a cumulative 90\% success rate in grasping the stalk. Overall, the robot autonomously deploys sensors into 29 of the 48 trial stalks, resulting in sensor insertion success rate of 60\%. 

The biggest cause of failure for this task is in not penetrating the sensor deep enough into the stalk. When a sensor is not deeply inserted, it may be knocked out of the stalk when the gripper retracts, as shown in insertion sequence step 6 in Fig. \ref{fig:sequence}. Such shallow insertions occur either when the sensor insertion direction deviates too far from the surface normal at the insertion point, or when the insertion point is offset from the vertical center of the stalk. When either of these deviations occur, the sensor often slides along the surface of the side of the stalk without penetrating, or penetrates only slightly. Contrary to our initial belief that cross-sections of cornstalks are mostly circular, stalk cross-sections are often elliptical in shape, and their growth orientation is stochastic in the field, making alignment difficult. For stalks with elliptical cross-sections, insertion positions along the minor axis achieve far greater success than those along the major axis, due to the less severe curvature.

\begin{figure}[t]
    \centering
    \includegraphics[width=\linewidth]{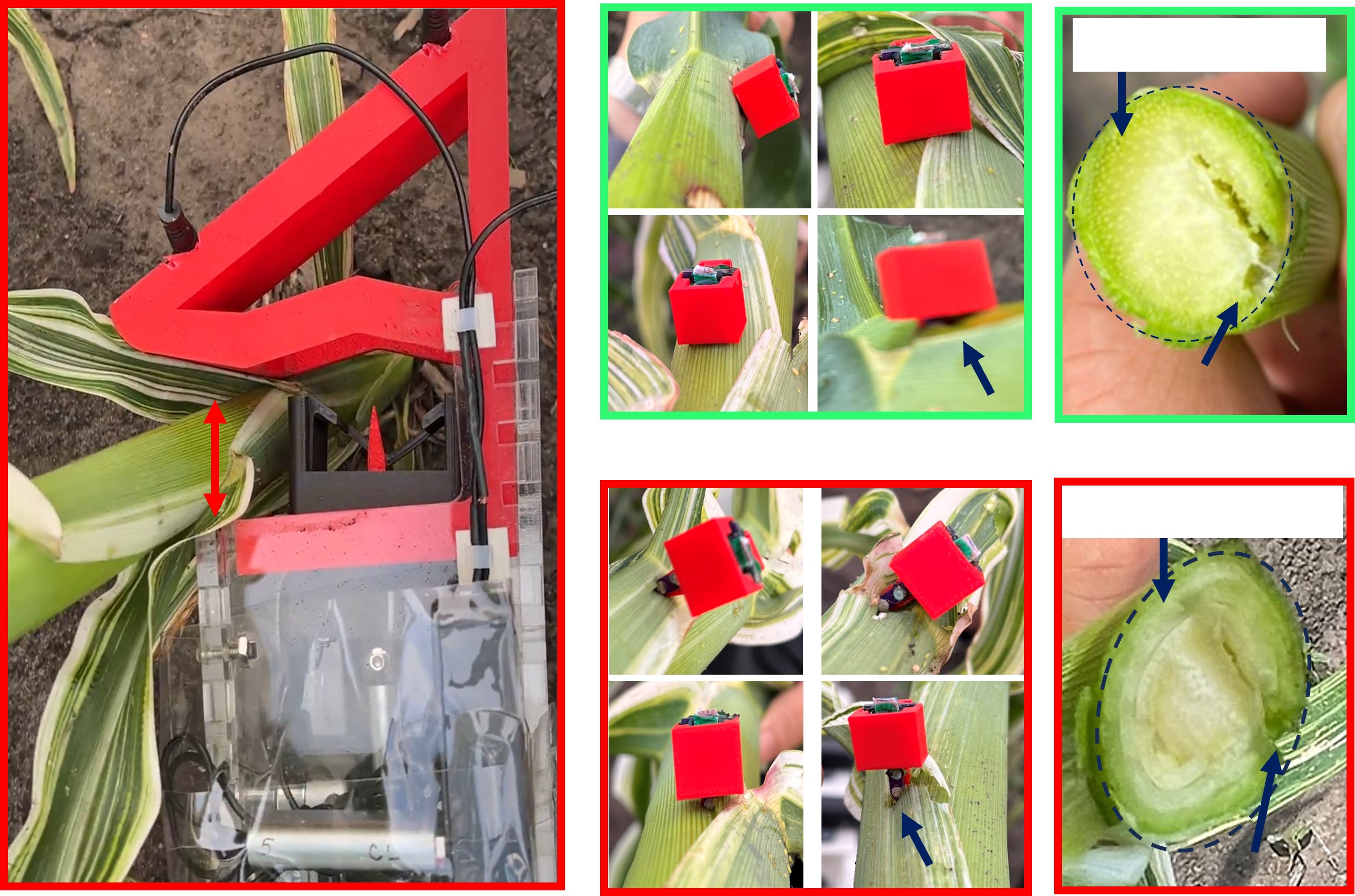}
    
    \setlength{\unitlength}{1cm}
    \begin{picture}(0,0)
    \put(-4.0,  3.4){\textcolor{red}{\scriptsize thickness}} 
    \put(-4.0,  3.2){\textcolor{red}{\scriptsize exceeds}} 

    \put(-4.0, 0.15){\footnotesize Failed to fit inside gripper}
    \put(-0.15,  0.15){\footnotesize Pads not covered} 
    \put(-0.15,  3.2){\footnotesize Pads fully covered} 
    \put(2.55,  0.15){\footnotesize Not through pith} 
    \put(2.55,  3.2){\footnotesize Through pith} 
    
    \put(2.75,  5.8){\scriptsize Solid filling} 
    \put(2.5,  2.8){\scriptsize Concentric layers} 
    
    \end{picture}
    \vspace{0pt}
    \caption{Visual examples of successful (green box) and failure cases (red box) of sensor insertion.}
    \label{fig:fail_visual}
    
\end{figure}
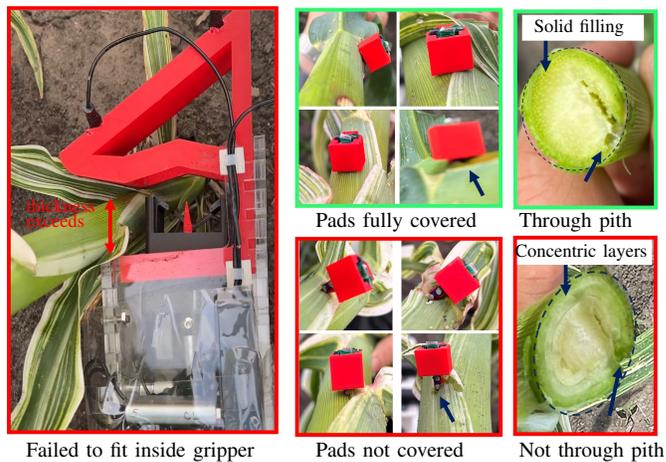

We conduct post-trial evaluation after each insertion attempt by visually inspecting that both electrode pads of the sensor are fully covered inside the plant. We also cut open each stalk to determine whether the sensor is inserted into the pith region. Of the 29 successful insertions, inspection reveals that in 16 trials, both sensor pads are sufficiently covered. Given the size of the sensor and approximate dimension of the stalks, even a 2 mm error in insertion depth often leads to an incomplete covering of the pads, as shown in Fig \ref{fig:fail_visual}. 
All except one of these trials have successful insertion in the pith region, which is verified by cutting open the stalk and exposing the pith region composed of a solid filling, as opposed to the undesired growth-region which is composed of layers of concentric structures that grow into leaves as shown in Fig \ref{fig:fail_visual}. 

\section{Discussion}~\label{sec:discussion} 
\vspace{-15pt}

In this work, we present a robot platform for agriculture research that can insert nitrate sensors into cornstalks. In this section, we discuss limitations and insights obtained from our experience in deployment to guide future developments in this field.

First, the manipulator design may be improved by accounting for more variation in stalk diameter and compliance. The funnel design of the V-sliding-block works best on young, compliant stalks and worse on more mature, rigid stalks. The funnel and V-sliding-block are designed such that when a misaligned stalk makes contact with the V-sliding-block, the stalk slides into the center of the funnel and aligns with the sensor. However, the fixed size of the funnel and fixed positioning of the V-sliding-block make the system less adaptable to larger and more rigid stalks, where there is less room for the V-sliding-block to align the stalk. In a future iteration, we suggest a mechanism which can fully envelope a stalk in its gripper and conform to its shape—in this way, the location of the stalk within the gripper mechanism can be more accurately known, resulting in more accurate insertions.

Second, the xArm robot motion sequence may be improved by further closing the perception-action loop and integrating sensor feedback from the manipulator to improve localization of a targeted stalk. While the demonstrated open-loop motion sequence with the gripper design works well, it cannot account for high variation in stalk diameter and rigidity. For example, the motion sequence remains the same for two stalks with different diameters. However, the distance the arm must retract (motion sequence step 4 in Fig. \ref{fig:sequence}) to align the sensor changes. Future work includes investigating additional sensing modalities, such as vibro-tactile feedback with an array of contact microphones \cite{tactile}, to reduce the uncertainly of the stalk position within the gripper. 

Lastly, the stalk detection system may be improved by accounting for stalk orientation and adding active lighting. As aforementioned, cornstalks with highly elliptical cross-sections are common and the current system disregards these shapes and returns a single insertion position; the system's success rate is also affected by the axis of the ellipse on which the insertion position lies. Thus, a perception pipeline which fully estimates a 3D ellipse at a given height would allow the robot arm to align itself with the most favorable insertion point. Additionally, active lighting within the perception system may alleviate the issue of poor model performance due to varying lighting conditions not captured in the dataset \cite{active_light}.
\vspace{-5pt}
\section{Conclusion}
\vspace{-5pt}
We demonstrate our robot platform’s capability of inserting sensors into cornstalks, and share design decisions and improvements needed for future research. The task of sub-centimeter-precision sensor insertion in highly cluttered and varying cornstalks is inherently difficult.
But understanding the real challenges in the field aids iteration on both hardware and software design to build a research platform that can autonomously insert sensors for phenotyping and crop management. We plan on  further improving the research platform for more scalable deployment of sensors, as well as incorporating navigation capabilities to achieve a fully autonomous platform.


 %



\section*{ACKNOWLEDGMENTS}
We would like to thank Vignesh Kumar, Prof. Liang Dong, and Prof. Patrick Schnable from Iowa State University for help at Curtiss farm. This work was  supported by: NSF Robust Intelligence 1956163, and NSF/USDA NIFA AIIRA AI Research Institute 2021-67021-35329.

\bibliographystyle{IEEEtran} 
\bibliography{mybib}
\end{document}